\pdfoutput=1

\documentclass[11pt]{article}

\usepackage{amsmath,amsfonts,bm}

\def\eqref#1{equation~\ref{#1}}

\def\1{\bm{1}}

\def\vh{{\bm{h}}}

\def\vx{{\bm{x}}}

\def\mA{{\bm{A}}}
\def\mB{{\bm{B}}}

\def\mM{{\bm{M}}}

\def\mS{{\bm{S}}}

\def\mW{{\bm{W}}}
\def\mX{{\bm{X}}}

\DeclareMathAlphabet{\mathsfit}{\encodingdefault}{\sfdefault}{m}{sl}
\SetMathAlphabet{\mathsfit}{bold}{\encodingdefault}{\sfdefault}{bx}{n}
\newcommand{\tens}[1]{\bm{\mathsfit{#1}}}

\def\tX{{\tens{X}}}

\newcommand{\R}{\mathbb{R}}

\usepackage[final]{acl}

\usepackage{times}
\usepackage{latexsym}

\usepackage[T1]{fontenc}

\usepackage[utf8]{inputenc}

\usepackage{microtype}

\usepackage{inconsolata}

\usepackage{booktabs}       
\usepackage{amsfonts}       
\usepackage{nicefrac}       
\usepackage{microtype}      

\usepackage{enumitem}
\usepackage{multirow}
\usepackage{adjustbox}
\usepackage{pifont}
\usepackage{colortbl}
\usepackage{nicefrac}       
\usepackage{microtype}  

\usepackage{algorithm}
\usepackage{algcompatible}
\usepackage{subcaption}
\usepackage{amsmath}
\usepackage{listings}
\usepackage{wrapfig}

\newcommand{\ourmethod}
{PRILoRA} 

\newcommand{\X}{X}

\newcommand{\Wqi}{\mW_{q_i}}
\newcommand{\Wki}{\mW_{k_i}}
\newcommand{\Wvi}{\mW_{v_i}}
\newcommand{\Wo}{\mW_{o}}

\newcommand{\bb}{\bm{b}}
\newcommand{\W}{W}

\newcommand{\bx}{\bm{x}}
\newcommand{\bh}{\bm{h}}

\title{{\ourmethod}: Pruned and Rank-Increasing Low-Rank Adaptation
}

\author{Nadav Benedek \\
  Tel Aviv University \\
  \texttt{nadavbenedek@mail.tau.ac.il} \\\And
  Lior Wolf \\
  Tel Aviv University \\
  \texttt{wolf@cs.tau.ac.il} \\}

\begin{document}

\maketitle

\begin{abstract}
With the proliferation of large pre-trained language models (PLMs), fine-tuning all model parameters becomes increasingly inefficient, particularly when dealing with numerous downstream tasks that entail substantial training and storage costs. Several approaches aimed at achieving parameter-efficient fine-tuning (PEFT) have been proposed. Among them, Low-Rank Adaptation (LoRA) stands out as an archetypal method, incorporating trainable rank decomposition matrices into each target module. Nevertheless, LoRA does not consider the varying importance of each layer. 
To address these challenges, we introduce {\ourmethod}, which linearly allocates a different rank for each layer, in an increasing manner, and performs pruning throughout the training process, considering both the temporary magnitude of weights and the accumulated statistics of the input to any given layer. We validate the effectiveness of {\ourmethod} through extensive experiments on eight GLUE benchmarks, setting a new state of the art.
\end{abstract}

\section{Introduction}

The current paradigm for natural language processing tasks is to exploit pre-trained models, which were trained using large amounts of data and expensive resources, and fine-tune them to various downstream tasks \citep{brown2020language, liu2019roberta,radford2019language, he2021deberta,devlin2018bert}. Such fine-tuning was traditionally conducted by gradient update of all parameters of the model \citep{dodge2020fine,raffel2020exploring, qiu2020pre}. With the ever increasing size of models, such as Llama 7B-65B \citep{touvron2023llama}, Palm 540B \citep{chowdhery2022palm}, and others, trained with resources consisting of hundreds of GPUs in parallel, which are available only to some institutions and corporations, full fine-tuning can become prohibitive, lengthy, and with high carbon footprint \citep{luccioni2022estimating}. Additionally, fully fine-tuning this way requires storing all parameters of the fine-tuned model for every downstream task.

To tackle the aforementioned challenges, a few research directions for Parameter-Efficient Fine-Tuning (PEFT) were proposed. These directions aim to maintain or even improve the accuracy of a full fine-tuning approach, while training only a small fraction of the parameters. One approach is to add small modules to the base model, which is kept frozen throughout the training process. Such adapter tuning techniques \citep{rebuffi2017learning,houlsby2019parameter,pfeiffer2020adapterfusion,he2022towards} add modules between the layers. The implication, due to increased model depth, is longer training time and higher latency during inference. Alternatively, prompt and prefix tuning \citep{lester2021power,li2021prefix} attach trainable tokens to the beginning of layers in the model, thus potentially reducing its effective maximal token length.

LoRA \citep{hu2022lora} fine-tunes linear layers by viewing each layer as a matrix of weights $W_0$, freezing it, and adding to it a small rank matrix, with the same shape as the original weight matrix, that is obtained as a product of two low-rank matrices $A$ and $B$. The low-rank $r$ is chosen to be much smaller than the input dimension to the layer, thereby significantly reducing the number of trainable parameters. During LoRA training, only the two low-rank matrices are updated, which are usually 0.01\% to 1.00\% of the original parameter count, depending on the low-rank of the two matrices. In addition to being efficient and often exceeding the performance of full fine-tuning \citep{hu2022lora}, this method has the advantage of being able to be merged back to the original matrix during inference, without increasing latency. LoRA has been used in various downstream tasks successfully \citep{schwartz2022maeday,lawton2023neural,dettmers2023qlora}

One limitation of LoRA is that the low-rank $r$ is an arbitrarily set parameter, and in the original LoRA it is set to be fixed across layers and weights. 

Efforts were made to address the issue of the fixed rank of LoRA. AdaLoRA \citep{zhang2023adaptive} starts from an initial parameter budget, which is slightly higher than the final budget, and then gradually reduces it until matching the target by removing weights based on SVD.

In this work, we encourage the usage of linearly increasing the rank from one layer to the next while concurrently adhering to the same budget of parameters. As we show, this strategy provides a distribution of the learned parameters that is better than a uniform placement, or even the learned alternatives.

A second contribution is obtained by pruning matrix $A$. This is done by considering both the elements of $A$ and an exponential moving average over the layer's input. Although we prune, in most cases, half of the elements of $A$, the main metric we seek to improve by pruning is the overall accuracy obtained after pruning. 

We conduct extensive experiments over eight different General Language Understanding Evaluation \citep{wang2018glue} benchmarks, and present evidence that the proposed method outperforms LoRA and its recent variants, that both the linear distribution of ranks and the specific pruning approach are beneficial, and that the method does not require more GPU memory or training time than the conventional LoRA, unlike recent extensions of LoRA.

\section{Related Work}

In recent years, Parameter Efficient Fine-Tuning (PEFT) has garnered increasing interest among researchers as a means to reduce both the expenses associated with fine-tuning and storing large-scale pre-trained models and the time required for training. Various approaches have emerged, each exhibiting distinct characteristics pertaining to memory utilization, storage requirements, and computational overhead during inference. These approaches can be classified into two primary categories, namely, selective and additive PEFT methods, based on whether the original model parameters undergo fine-tuning during the training phase.

\textbf{Selective methods} involve the selection and modification of a model based on its original parameters. An early instance of this concept was observed in the fine-tuning of only a subset of the top layers of a network, as demonstrated by \citet{donahue2014decaf}, and by more recent work \citep{gheini2021cross}. In more recent developments, various approaches have been proposed, each targeting specific layers or internal modules of the model. For instance, the BitFit method \citep{zaken2021bitfit} updates only the bias parameters, resulting in a substantial reduction in the number of trainable parameters, but at the cost of suboptimal performance. Other methods use a scoring function when selecting trainable parameters \citep{guo2020parameter,sung2021training,vucetic2022efficient}, while others select top parameters based on a Fisher information calculation \citep{sung2021training}.

\textbf{Additive methods} represent an alternative to full-parameter fine-tuning by introducing additional trainable parameters into the backbone network. Adapters are a type of trainable component initially applied in the context of multi-domain image categorization by \citet{rebuffi2017learning}, that were subsequently integrated into Transformer networks, specifically in the attention and feed-forward layers \citep{houlsby2019parameter}. Prefix-Tuning and Prompt-Tuning \citep{li2021prefix,lester2021power} involve the addition of trainable parameters preceding the sequence of hidden states across all layers. LST (Ladder Side-Tuning) \citep{sung2022lst} operates by short-cutting hidden states from the original network into a compact trainable side network, eliminating the need for backpropagating gradients through the backbone network.

LoRA \citep{hu2022lora} emulates the adjustment of the weight matrix in the model through the multiplication of two low-rank matrices. Notably, the trained parameters resulting from this process can be incorporated seamlessly into the original network during the inference phase without incurring additional computational overhead.

Recently, hybrid approaches have emerged, combining the selective and additive methods and presenting a unified framework \citep{chen2023parameter,he2022towards,mao2021unipelt}. Other methods are based on the hypothesis that parameter redundancy exists in PEFT modules, therefore pruning the trainable parameters to achieve superior
 fine-tuning performance \citep{bai2022parameter}.

{\bf Network pruning} methods \citep{molchanov2016pruning,hassibi1993optimal,frankle2018lottery,liu2018rethinking,han2015learning} reduce the size of the network by removing or shrinking matrices from the network, which effectively is equivalent to setting them to zero. Such methods require further full re-training, or other computationally intensive iterations. 

{\bf{Magnitude Pruning}} \citep{han2015deep,gale2019state} removes individual parameter weights when the magnitude is below a certain threshold. The threshold is determined either based on the relative magnitude to other weights in the same parameter or layer \citep{zhu2017prune}, or for the whole network \citep{liu2018rethinking}.

\section{Background}

{\bf Transformer Models. } 
Transformer~\citep{vaswani2017attention} is a sequence-to-sequence architecture that makes use of self-attention. Typically, it consists of several stacked blocks, where each block contains two sub-modules: a multi-head attention (MultiHead) and a fully connected feed-forward network (FFN). Given the input sequence $ \mX \in \R^{n\times d} $ of $n$ tokens of dimension $d$, MultiHead performs the attention function using $ h $ heads, allowing each segment of the $d$ space to attend to a different value projection of another token: 

\begin{align*}
    \text{MultiHead}\left( \mX \right) = [\text{head}_1,.., \text{head}_h]\mW_o  \in \R^{n\times d} 
\end{align*}
\begin{align*}
	\text{head}_i = \text{Softmax}\left( \frac{{\mX \Wqi (\mX\Wki)^{\top} }}{{\sqrt{d_h}} }  \right) (\mX\Wvi)
\end{align*}
where the square brackets denote a concatenation along the second dimension, $ \Wo\in\R^{d\times d} $ and $ \Wqi,\Wki,\Wvi \in\R^{d\times d_h} $ are parameters of head $ i $, per block, and the softmax is applied to each row.  $ d_h $ is typically set to $ \frac{d}{h} $. The output of the MultiHead is fed into the FFN, consisting of two linear transformations with a ReLU non-linearity in between: 

$\text{FFN}(\X) = \text{ReLU}(\mX\mW_1 + \bb_1)\mW_2+ \bb_2$,  where $ \mW_1 \in\R^{d\times d_m} $ and $ \mW_2 \in \R^{d_m \times d} $ are parameters of the block. Lastly, a residual connection is applied and a layer normalization \citep{ba2016layer}. \\

\noindent{\bf Adapters.\quad} \citep{houlsby2019parameter,pfeiffer2020adapterfusion} The adapter technique injects a module between the transformer layers, such that the input is down-projected to a lower-dimensional space using $\mW_{down} \in \R^{d \times r}$, followed by non-linearity $\sigma$, and up-projected using $\mW_{up} \in \R^{r \times d}$, combined with a residual connection:
\begin{align}
\vh=\vx+\sigma(\vx \mW_{down})\mW_{up}
\end{align}
\\
{\bf Low Rank Adaptation. }
LoRA \citep{hu2022lora} freezes the pre-trained model weights and injects two trainable rank decomposition matrices into each
layer of the Transformer architecture, greatly reducing the number of trainable parameters for fine-tuning tasks. For a linear layer $ \bh=\W_0 \bx $, the LoRA-modified forward function is: 
\begin{align}
\label{eq:forward_lora}
\vh = \mW_0 \vx + \Delta \mW \vx = \mW_0 \vx + \mB\mA \vx 
\end{align}
where  $ \mW_0,\Delta \mW \in\R^{d_1\times d_2} $, $ \mA\in\R^{r \times d_2} $ and $ \mB\in \R^{d_1\times r} $ with $r\ll \{ d_1, d_2 \} $. $ \mA $ is Gaussian initialized and $ \mB $ is zero initialized, in order to have $ \Delta \mW = 0 $ at the beginning of the fine-tuning training. 
\citet{hu2022lora} apply LoRA to the query and value parameters (i.e,~$ \mW_q $ and $ \mW_v $) in the multi-head attention, without modifying the other weights. \citet{he2022towards} extend it to other weight matrices of the feed-forward network, for an increased performance. 

\section{Method}

Our proposed method, {\ourmethod} (Pruned and Rank-Increasing Low-Rank Adaptation), is comprised of two main components that integrate with the LoRA fine-tuning: (i) Linear distribution of low ranks across the layers in the network, and (ii) Ongoing pruning of the $\mA$ matrix of the LoRA, based on the layer's input activations and the weights of the LoRA $\mA$ matrix.

\subsection{Linear Distribution of Ranks}

While LoRA distributes the learned parameters uniformly, one can distribute these differently. For example, one can assign a lower rank to some of the layers and a higher rank to others. 

Recall that the trainable parameters in LoRA are the matrices $\mA$ and $\mB$. Each has one dimension that is fixed according to the layer's structure, and one dimension that is the low rank $r$. Since both the time complexity (train or test) and the memory complexity of a layer are linear in both the input and the output dimensions of each layer, and since only one dimension of $\mA$ and $\mB$ depends on $r$, the overall complexity of LoRA is linearly dependent on the sum of the ranks in all modified layers.

The way that we distribute the learned parameters is motivated by the results provided by~\cite{zhang2023adaptive}, which demonstrate that the top layers require more adaptation. Considering that one cannot focus only on the top layers, since the other layers also need to adapt (see Sec.~\ref{sec:discussion}), and to promote simplicity, we employ a linear distribution of ranks. 

In the linear distribution of ranks, we allocate a different low-rank for every layer in the model, in a linearly increasing manner. Specifically, for the DeBERTaV3-base model, we start from the first layer, applying a low-rank of $r_s=4$, and growing linearly, up to the twelfth layer, where we apply $r_f=12$, such that the average rank across layers is 8. We allocate the same low-rank to all weights in a given layer, regardless of the matrix type (query, key, value, etc.). This makes the total number of parameters identical to the LoRA method.  

\subsection{Ongoing Importance-Based A-weight Pruning}

We employ pruning as a form of dynamic feature selection, which allows the fine-tuning process to focus on some of the layer's input at each bottleneck index at every pruning iteration. The intuition is that since the capacity of the update matrix $\mB \mA$ is low, it would be beneficial to attend only to the important input dimensions. 

\subsubsection{Importance Matrix}

\begin{figure}
\centering
    \includegraphics[width=.807\linewidth]{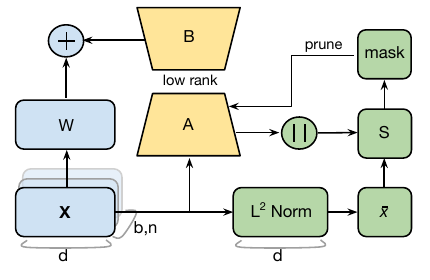}
    \captionof{figure}{The schematics of {\ourmethod} on a single layer. The blue path demonstrates a frozen linear layer. We omitted the bias for simplicity. The yellow path depicts LoRA; dropout and scaling were omitted for simplicity. In the green path of {\ourmethod}, the input tensor $\tX$ of the layer is fed into $L2$ norm calculation. Then, the exponential moving average vector $\bar \vx$ is updated and kept as a state of the layer. When it is time for pruning, the absolute value of the elements of $\mA$ is calculated, and together with $\bar \vx$, the importance matrix $\mS$ is computed. In every row of $\mS$, the lowest elements, as defined by the \textit{prune ratio}, are being selected to form the mask. The mask is used to zero out elements in the $\mA$ matrix.}
    \label{fig:prune_method}
\end{figure}

Each transformer layer, whether it is a projection associated with key, query, or value, or one of the FFN layers has some weight matrix $\mW$. It also has some input $ \tX \in \R^{b\times n\times d} $, where $b$ is the batch size, $n$ is the number of tokens, and $d$ is the dimension. We abuse the notation slightly and also write $ \tX $ for the second layer of the FFN, although, in this case, the dimension is $d_m$, which is typically larger than $d$. In our framework we maintain, throughout the training process, an Exponential Moving Average of the $L2$ norm of the rows of each such input $\tX$, as depicted in Figure~\ref{fig:prune_method}.

For each batch, we consider the tensor that has a dimension of $b\times n \times d$, square all elements, sum across the first and second dimensions, obtaining a vector of size $d$, and take the square root of each vector element, to get $\vx$. 

The exponential moving average $\bar \vx$ is updated between batches by the following rule
\begin{equation}
    \bar \vx = 0.9 \bar \vx + 0.1 \vx
\end{equation}

We next compute, for every weight matrix $\mW$, or, more specifically, for $\mA\in \mathbb{R}^{r\times d_2}$, which is the associated half-decomposition of $\Delta \mW$, an importance matrix $\mS$ of the same size as $\mA$. $\mS$ is inspired by Wanda~\citep{sun2023simple}, and is the element-wise multiplication of the absolute value of $\mA$ with the relevant moving average vector $\bar \vx$ (recall that there is one $\bar \vx$ to each weight matrix $\mW$):
\begin{align}
	 \mS_{ij} = |\mA_{ij}| {\bar \vx_j}
\end{align}

Note that all values of $\bar \vx$ are positive, since they represent a mean norm. Therefore, all elements of $\mS$ are positive, too.

\begin{figure*}[t]
      \centering
      \begin{tabular}{cc}
    \includegraphics[width=.494945\linewidth]{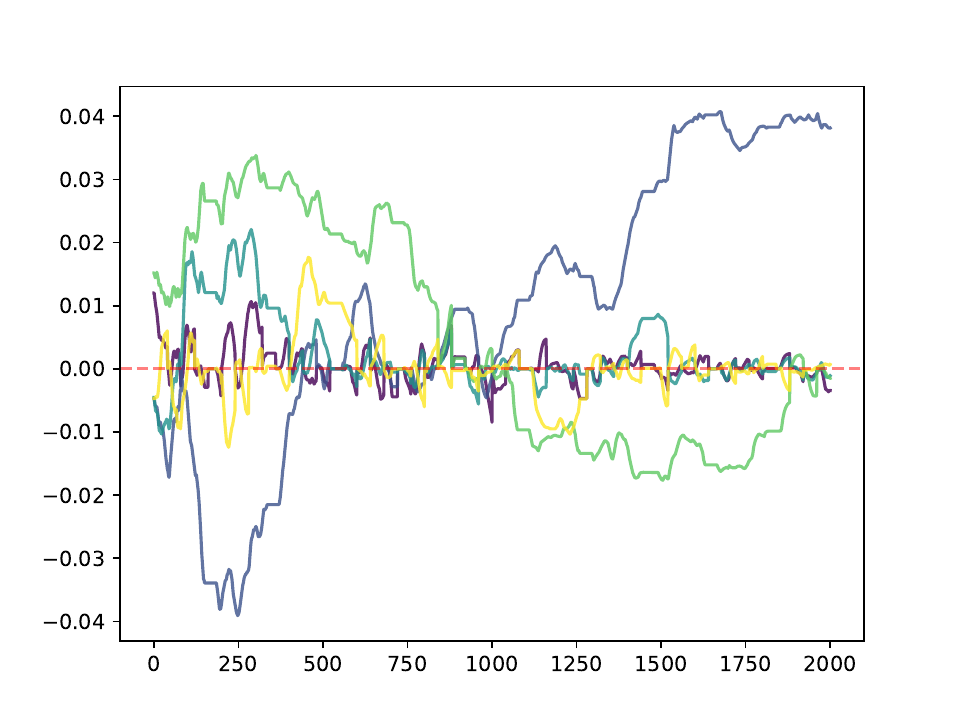}&
    \includegraphics[width=.494945\linewidth]{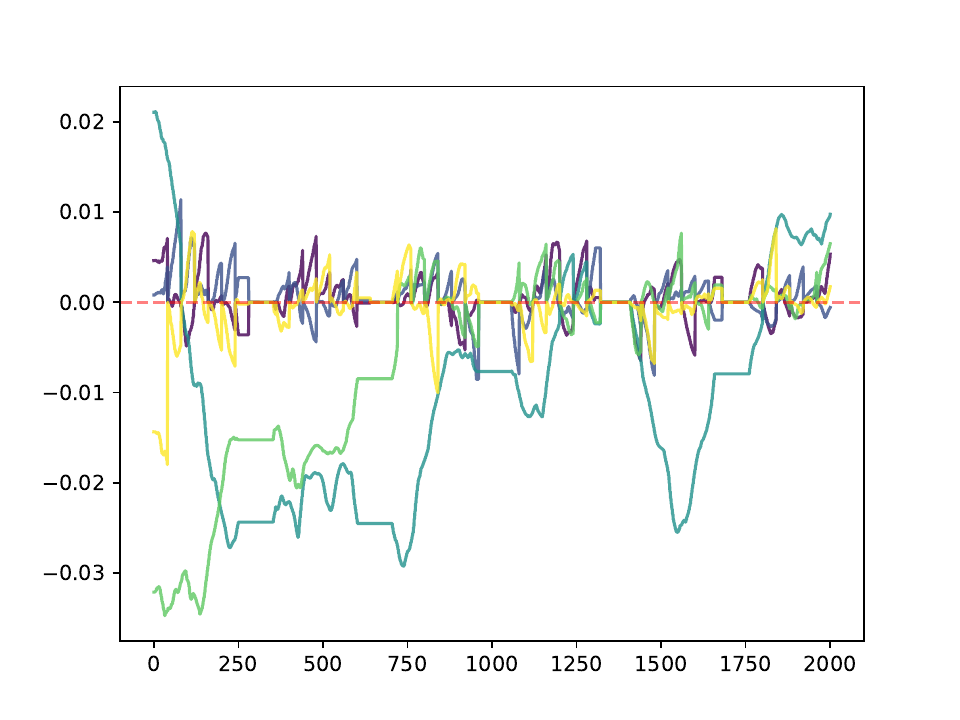}\\
        (a) &(b)\\
    \includegraphics[width=.494945\linewidth]{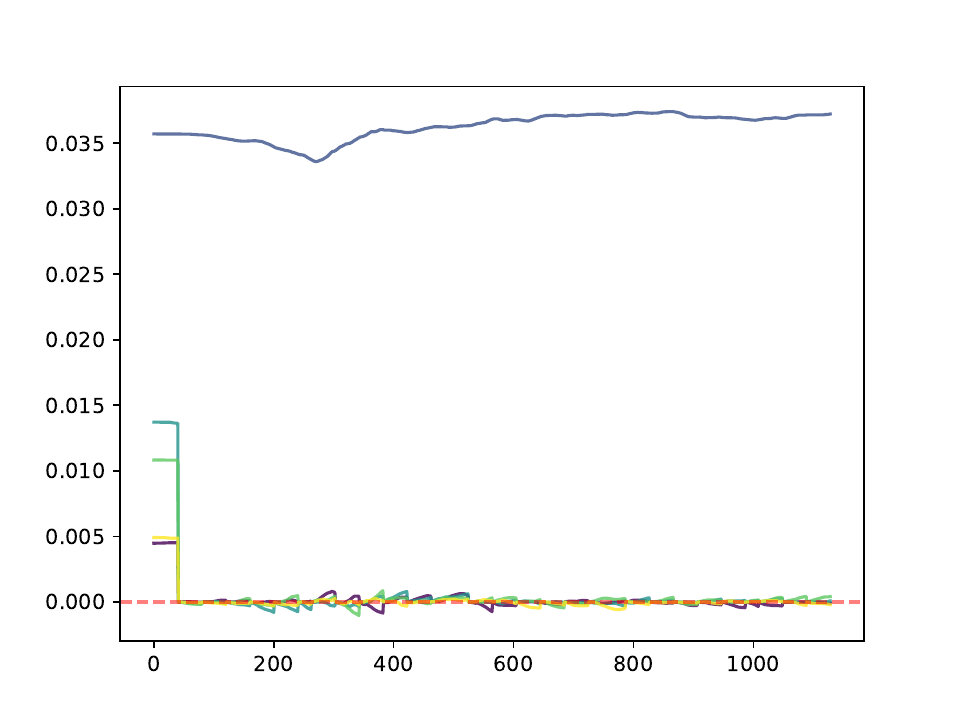}&
    \includegraphics[width=.494945\linewidth]{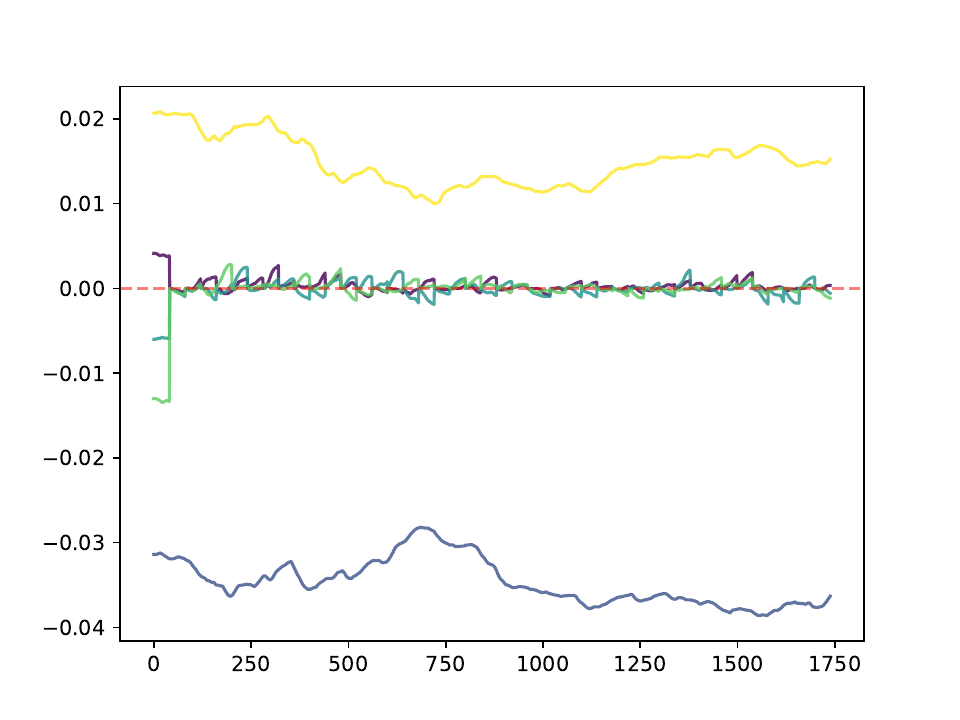}\\
    (c) & (d)\\
    \end{tabular}
    \caption{Five weights values over time on four different GLUE tasks: (a) RTE task, in layer 5, value\string_proj parameter; (b) MRPC task, in layer 6, query\string_proj parameter; (c) SST-2 task, in layer 7, key\string_proj; (d) CoLA task, in layer 8, attention.output parameter.}
    \label{fig:prune_graph}
 \end{figure*}

\subsubsection{Pruning}

 Every 40 steps in the training process, we prune each of the $\mA$-matrices, in accordance with the associated importance matrix $\mS$. To do so, we consider the $n$ lowest elements of every row $i=1\dots r$ of $\mS$ and create a binary mask $\mM\in \mathbb{R}^{r\times d_2}$. Each mask element $\mM_{ij}$ indicates whether $\mS_{ij}$ is among the $n$ lowest values of row $i$ of $\mS$. $n$ is determined by the \textit{prune ratio}; a higher ratio means more weights are being zeroed out. We then zero out the elements in $\mA$ using the mask $\mM$.

Note that zeroing out an element of $\mA$ does not prevent this element from becoming non-zero immediately in the next training step. However, pruning this way changes the training dynamics and encourages $\mA$ to be sparse. Figure~\ref{fig:prune_graph} shows five random weights during training of different datasets. It can be seen that some weights can survive pruning, some weights remain in the pruning region since they cannot escape fast enough, and some weights avoid being pruned completely.

\section{Experiments}

We apply {\ourmethod} to DeBERTaV3-base \citep{he2021debertav3} (184 million parameters), and evaluate the method on eight natural language understanding benchmarks included in the General Language Understanding Evaluation - GLUE \citep{wang2018glue}. Summary of the GLUE benchmarks can be found in Table~\ref{tab:glue_dataset}. We use PyTorch \citep{paszke2019pytorch} and Hugging Face Transformers \citep{wolf2019huggingface} to implement the algorithms. All the experiments are conducted on NVIDIA GeForce RTX 2080 Ti GPUs. Due to limited GPU memory size, we leave similar analysis of large-scale models, such as T5-3B, Llama, and others, to future research. 

\subsection{Baselines}

\textbf{Full fine-tuning}: In the fine-tuning stage, the model is initialized with the pre-trained parameters, and all model parameters go through gradient updates.

\textbf{Bitfit}: \citep{zaken2021bitfit} A sparse fine-tuning method where only the bias-terms of the model (or a subset of them) are being modified.

\textbf{HAdapter}: \citep{houlsby2019parameter} Inserts adapter layers between the self-attention module, the FFN module, and the subsequent residual connection. There are two
fully connected layers with biases in an adapter layer with a non-linearity in between.

\textbf{PAdapter}: \citep{pfeiffer2020adapterfusion} Inserts the adapter after
the FNN module and LayerNorm.

\textbf{LoRA}: \citep{hu2022lora} Adds trainable pairs of rank decomposition matrices in parallel to existing weight matrices. The number of trainable parameters is determined by the rank $r$ and the shape of the original parameters.

\textbf{AdaLoRA}: \citep{zhang2023adaptive} Parameterizes the incremental
updates in the form of singular value decomposition, for a given parameter.

\subsection{Implementation details}
In our research, we experimented with different distributions while keeping the total number of parameters invariant and found that the configuration $\{ r_s=4, r_f=12 \}$ was optimal, together with the hyper-parameters which are specified in Table~\ref{tab:glue_hypers}.
The fact that higher layers require more parameters for LoRA fine-tuning may indicate that higher layers in Transfomer-based models capture deeper levels of understanding, and therefore when fine-tuning a pre-trained language model, more focus must be put on deeper layers than on lower layers that require less modification or adaptation to the downstream task in question.

\subsection{Main results}

We compare {\ourmethod} with the baseline methods. Table~\ref{tab:glue_datasets} shows our results on the GLUE development set  (Appendix~\ref{app:glue_datesets}). {\ourmethod} achieves best average score, best result in six out of the eight datasets, and in all datasets better results than HAdapter, PAdapter and LoRA, with approximately the same number of parameters. 

Note that when counting the number of parameters, we do not discount for pruned parameters. However, with a pruning ratio of 0.5 in most benchmarks, a quarter of the learned parameters (half the parameters of the $A$ matrices) are zero. A more precise count of parameters would, therefore, be closer to one million parameters and not 1.33M.

{\setlength{\tabcolsep}{0.35em}
\renewcommand{\arraystretch}{1.1}

\begin{table*}[t]
\vspace{3mm}
\caption{Results with DeBERTaV3-base on GLUE development set. The best results on each dataset are shown in \textbf{bold}. 
We report the average correlation for STS-B (Pearson, Spearman). We report matched accuracy for MNLI. \textit{Full FT}, \textit{HAdapter} and \textit{PAdapter} represent full fine-tuning, Houlsby adapter, and Pfeiffer adapter, respectively. 
We report the mean and standard deviation of three runs using different random seeds. We report the baseline results from \citet{zhang2023adaptive}. Higher is better for all metrics.}

\vspace{1mm}
\begin{center}

\begin{tabular}{lcccccccccc}
\toprule
\multirow{2}*{\bf Method} & \multirow{2}*{\bf \small \#Param} & {\bf MNLI} & {\bf SST-2} & {\bf CoLA} & {\bf QQP } & {\bf QNLI} & {\bf RTE}  & {\bf MRPC}  & {\bf STS-B} & {\bf All} \\
~ & ~ & {Acc} & {Acc} & {Mcc} & {Acc} & {Acc} & {Acc} & {Acc} & { Corr } & {Avg.} \\
\midrule 
{Full FT} & {184M} & {89.90} & {95.63} & {69.19} & {92.40} & {94.03} & {83.75} & {89.46} & {91.60} & {88.25}
\\
\midrule
{BitFit} & {0.1M} & {89.37} & {94.84} & {66.96} & {88.41} & {92.24} & {78.70} & {87.75} & {91.35} & {86.20}  
\\
\midrule 
{\small HAdapter} & {1.22M} & {90.13} & {95.53} & {68.64} & {91.91} &  {94.11} & {84.48} & {89.95} & {91.48} & {88.28} 
\\
{\small PAdapter} & {1.18M} & {90.33} & {95.61} & {68.77} & {92.04} & {94.29} & {85.20} & {89.46} & {91.54} & {88.41}  
\\
{$\text{LoRA}_{r=8}$} & {1.33M} & {90.65} & {94.95} & {69.82} & {91.99} & {93.87} & {85.20} & {89.95} & {91.60} & {88.50} 
\\
AdaLoRA & {1.27M} & {\bf 90.76} & {96.10} & {71.45} & {92.23} & {\bf94.55} & {88.09} & {90.69} & {91.84} & {89.46} 
\\
{\ourmethod} & {1.33M} & {90.75} & {\bf 96.21} & {\bf72.79} & {\bf92.45} & {94.44} & {\bf 89.05} & {\bf92.49} & {\bf91.92} & {\bf90.01} 
\\
\multicolumn{2}{r}{\footnotesize[\ourmethod~ SD]}  & {$\pm$0.03} & {$\pm$0.30} & {$\pm$1.28} & {$\pm$0.05} & {$\pm$0.14} & {$\pm$1.04} & {$\pm$0.57} & {$\pm$0.14} & {$\pm$0.44}\\

\bottomrule
\end{tabular}
\end{center}
\label{tab:glue_datasets}
\end{table*}
}
\begin{table}[t]
\vspace{2mm}
\caption{Ablation study results on the same single seed.} 
\vspace{-1mm}
\label{tab:diplora_ablation}
\begin{center}
\begin{tabular}{@{}l@{~}c@{~~}c@{~~}c@{~~}c@{}}
\toprule
~ & SST-2 & CoLA & RTE &  MRPC
\\
\midrule 
{{\ourmethod}} & \bf96.44 & \bf73.08 & \bf90.25 & \bf93.14
\\
\midrule
{Fixed distribution} & 96.10 & 72.17 & 88.81 & 92.16
\\
{Inverted distribution} & 95.99 & 69.73 & 88.09 & 91.91
\\
{Concentrated dist.} & 95.07 & 69.92 & 87.73 & 89.95
\\
\midrule
{No pruning} & 96.22 & 71.31 & 89.89 & 92.09
\\
{Prune B rows} & 96.10 & 71.41 & 89.89 & 91.67
\\
{Prune B cols.} & 96.22 & 71.46 & 88.81 & 91.42
\\
{Prune A rand cols.} & 94.84 & 70.75 & 88.09 & 89.22
\\
\bottomrule
\end{tabular}
\end{center}
\end{table}

\begin{table*}[t]
\vspace{2mm}
\caption{Performance vs Pruning Ratio. Each cell in the table shows the average across three different seeds, together with the standard deviation.}
\vspace{-1mm}
\label{tab:diplora_prune_ratio}
\begin{center}
\begin{tabular}{lcccccccc}
\toprule
~ & SST-2 & CoLA & RTE &  MRPC
\\
\midrule 
{Prune 0.25} & 96.10 $\pm$ 0.34 & 71.43 $\pm$ 0.30 & 87.73 $\pm$ 1.25 & 91.34 $\pm$ 0.99
\\
{Prune 0.50} & \bf96.21 $\pm$ 0.30 & \bf72.79 $\pm$ 1.28 & \bf89.05 $\pm$ 1.04 & \bf92.49 
 $\pm$ 0.57
\\
{Prune 0.75} & 95.95 $\pm$ 0.17 & 70.63 $\pm$ 1.56 & 87.73 $\pm$ 0.73& 90.85 $\pm$ 0.51
\\

\bottomrule
\end{tabular}
\end{center}

\end{table*}
\begin{table}[t]
\vspace{2mm} 
\caption{Comparison of memory consumption and time per epoch in training,  between {\ourmethod} and LoRA on NVIDIA GeForce RTX 2080 Ti GPU, with a batch size of 32. All models have  1.33M parameters.} 
\vspace{-2mm}
\label{tab:training_cost}
\begin{center}
\begin{tabular}{@{}c@{~~}ccc@{}}
\toprule
{ Dataset} &  { Method} & { GPU Mem} & { Time/epoch}
\\
\midrule 
\multirow{2}*{MNLI} 

~ &  {LoRA} & 9.559 GB & 117 min
\\
~ &   {\ourmethod} & 9.559 GB & 120 min  
\\

\midrule 

\multirow{2}*{SST-2} 

~ &  {LoRA} & 9.559 GB & 24 min
\\
~ &  {\ourmethod} & 9.559 GB & 23 min  
\\

\midrule 

\multirow{2}*{ QQP} 

~ &  {LoRA} & 9.559 GB & 109 min
\\
~ &  {\ourmethod} & 9.559 GB & 110 min  
\\

\bottomrule
\end{tabular}
\end{center}
\end{table}
\begin{table}[t]
\vspace{2mm}
\caption{Number of steps to evaluation peak point, on four selected GLUE tasks.} 
\vspace{-1mm}
\label{tab:diplora_steps}
\begin{center}
\begin{tabular}{lcccccccc}
\toprule
~ & SST-2 & CoLA & RTE &  MRPC
\\
\midrule 
{{\ourmethod}} & 9875 & 12375 & 1875 & 1750
\\
{LoRA} & 6500 & 8000 & 3250 & 1250
\\

\bottomrule
\end{tabular}
\end{center}
\end{table}

\subsubsection{Ablation Study}
In table~\ref{tab:diplora_ablation} we present an ablation study for {\ourmethod}, on four GLUE tasks: SST-2, CoLA, RTE and MRPC. We aim to analyze both the rank distribution across layers and the pruning method.

For the rank distribution study we: (i) remove the linear distribution component of our method, retaining the pruning component alone with identical rank at each layer; (ii) replace the 4$\xrightarrow{}$12 distribution by 12$\xrightarrow{}$4; (iii) attach LoRA adapter to only the last layer, with a higher rank of 24 (Concentrated Distribution).

For the pruning method study we: (i) remove the importance pruning component, retaining increasing rank distribution 4 $\xrightarrow{}$ 12; (ii) prune the rows of $\mB$ matrix instead of $\mA$, by collecting an exponential moving average of $\mB$ input norm, instead of the input to $\mA$ (or the layer); (iii) similarly, prune $\mB$ columns instead of rows; (iv) prune the columns of $\mA$ randomly, instead of {\ourmethod} method, but with the same \textit{prune ratio}.
During all ablation tests, per benchmark, we keep the same hyper-parameters and change only a single component. For all cells in the table, the same single seed is used.

\paragraph{Rank Distribution}

As can be seen, removing the linear distribution of the low-rank and fixing a constant rank across all layers, such that the total number of parameters stays the same as in LoRA, but applying pruning, reduces the results in all tests. {Removing the linear distribution nonetheless outperforms LoRA results, signalling that pruning is indeed an essential component of the method. For example, {\ourmethod} with no linear distribution on the SST-2 benchmark reaches 96.10, while LoRA is 94.95, and on CoLA it is 72.17 versus 69.82.}

Interestingly, changing the order of the rank allocation, to be 12$\xrightarrow{}$4, reduces the performance significantly; for example, a decrease of 73.08 $\xrightarrow{}$ 69.73 on the CoLA benchmark, and 93.14 $\xrightarrow{}$ 91.91 on the MRPC benchmark. Inverting the rank allocation order diminishes performance below fixed-rank allocation across layers. This provides additional support in the need to allocate more parameters to the top layers. 

Lastly, attaching LoRA only to the last layer yields the lowest average results across the rank distribution ablation study, for example 89.95 versus 93.14 on MRPC when the full method is used.

\paragraph{Pruning Method}

{Ablating pruning completely, reduces the performance. For instance, on CoLA it is reduced 73.08 $\xrightarrow{}$ 71.31. This is higher than LoRA (69.82), pointing to the positive effect of the rank-increasing distribution.}
{When we prune matrix $\mB$ instead of $\mA$, we obtain results similar to no pruning at all, suggesting that pruning $\mB$ did not yield any discernible benefits.}

{A plausible argument is that the input activation shape of $\mA$ and $\mB$ is very different, for example 768 versus 8, in the case of most weights in DeBERTaV3-base model, and a low-rank of 8. Choosing to row-prune matrix $\mB$ with a \textit{prune ratio} of 0.5, essentially means eliminating 4 out of 8 cells in every $\mB$ row, which can be too aggressive. Additionally, doing the same process on $\mB$ columns can create situations where a complete row of $\mB$ is zeroed out, which means that the corresponding output cell of LoRA will be zero as well. Furthermore, the compressed low-rank latent input to matrix $\mB$ already encapsulates the essential information, so pruning it deteriorates the performance.}

Finally, performing a random pruning of columns in $\mA$ with the same \textit{prune ratio}, produces the lowest results in the Pruning Method ablation study. 

\subsubsection{Pruning Ratio Study for \ourmethod}

We would like to learn how aggressive pruning should be, that is, how much sparsity should be injected into the LoRA weights in order to reach peak performance. We chose four GLUE tasks, and for each task and for each \textit{prune ratio} in \{0.25, 0.50, 0.75\} we ran the fine-tuning three times, each time with a different seed. We report the average result and standard deviation across the different seeds.

Table~\ref{tab:diplora_prune_ratio} shows that for the selected tasks, the optimal pruning ratio is 0.5. However, specifically for the STS-B task, a random hyper-parameter search yielded an optimal pruning ratio of 0.75, as can be seen in Table~\ref{tab:glue_hypers}.

\subsubsection{Training Cost Study for \ourmethod}

We present the training cost comparison between {\ourmethod} and LoRA, using the DeBERTaV3-base model, on NVIDIA GeForce RTX 2080 Ti GPUs. For the two methods, the batch size is 32.

Table~\ref{tab:training_cost} shows that {\ourmethod} has zero increase in number of trainable parameters in comparison to LoRA, and a negligible increase in training time per epoch.

For comparison, AdaLoRA \citep{zhang2023adaptive} speed per batch is 11\% slower than LoRA in the MNLI benchmark and 16\% slower in the SST-2 benchmark, and with a slightly larger memory footprint.

However, analyzing the training time per batch does not suffice. Once we know that the training step time in {\ourmethod} is similar to LoRA, we want to delve deeper and analyze the number of steps required until reaching peak performance on the evaluation metric.

Table~\ref{tab:diplora_steps} presents the number of steps required for each method until reaching its peak evaluation performance. Evidently, there is no clear winner with respect to the number of steps or time required to reach peak performance. Both LoRA and {\ourmethod} have the same order of magnitude. Since one often trains beyond the peak point, the table does not indicate that one method is preferable to the other in this respect.

\section{Discussion}
\label{sec:discussion}

Moving from one task to another requires an adaptation of both the input and the output domain. While the input domain of large language models may be comprehensive enough to support new downstream tasks, the generation of the output is very much context-and-task-dependent.

Therefore, it should not come as a surprise that fine-tuning requires more adaptation of the top layers, which are closer to the output, than of the earlier, input-processing, layers.

However, if one is to change only the top layers, as we showed in the ablation study, there would not be enough co-adaptation of the earlier layers to enable the top layers to produce the required output. It seems, therefore, that the gradual increase in the allocated resources, which we apply, is a reasonable strategy. 

\section{Conclusions}

In this paper, we introduced {\ourmethod}, a novel, yet simple and parameter-efficient method for improving low-rank adaptation during fine-tuning. Our extensive experiments encompass eight GLUE benchmarks across multiple seeds, illustrating the effectiveness of {\ourmethod}. Notably, we achieve superior performance compared to state-of-the-art metrics while maintaining the same number of trainable parameters, reducing the non-zero parameters by a quarter on most benchmarks, and adhering to the same memory constraints and running time per epoch.

\section{Limitations}

Our work has some limitations. We pushed the limits of our computational resources, utilizing NVIDIA GeForce RTX 2080 Ti GPUs, to conduct the experiments presented in this study across the eight GLUE benchmarks. We employed the {\ourmethod}-modified DeBERTaV3-base model, which consists of 184 million parameters. 

These experiments are of the same scale as the most related work~\citep{zhang2023adaptive}. However, the full potential of the method could be realized on larger models trained on more extensive datasets, and by using larger batches that can fit into GPU memory, allowing examination of the method on additional downstream tasks, such as question answering and text summarization. 

\clearpage

\bibliography{diplora,anthology,custom}

\appendix

\section{GLUE Dataset}\label{app:glue_datesets} 

Here is a summary of the benchmarks and metrics we used from the GLUE \citep{wang2018glue} dataset.

\begin{table*}[b]
	\begin{center}
	\caption{Summary of the GLUE dataset}
	\label{tab:glue_dataset}
		\begin{tabular}{l|l|c|c|c|c}
			\toprule 
			\bf Corpus &Task& \#Train & \#Dev & \#Label &Metrics\\ 
   
            \midrule
            
			\multicolumn{4}{@{\hskip1pt}r@{\hskip1pt}}
   {Single-Sentence Tasks} \\             
            \hline
		
   CoLA & Grammatical Acceptability&8.5k & 1k & 2 & Matthews corr\\ \hline
			SST-2 & Sentiment&67.3k & 872 & 2 & Accuracy\\ 
   
            \midrule
   
			\multicolumn{4}{@{\hskip1pt}r@{\hskip1pt}}{Pairwise Text Tasks} \\ \hline
   
			MNLI & NLI (Entailment) & 392k& 9.8k & 3 & Matched Accuracy\\ \hline
			RTE & NLI (Entailment) &2.5k & 277 & 2 & Accuracy \\ \hline
			
			QQP & Semantic Equivalence &364k & 40k & 2 & Accuracy\\ \hline
			MRPC & Semantic Equivalence &3.7k & 408 & 2&Accuracy\\ \hline
			QNLI & Question Answering& 105k &5.5k&2& Accuracy \\ \hline
			STS-B & Similarity & 5.7k &1.5k &1 & Pearson/Spearman corr\\ \bottomrule
			
		\end{tabular}
	\end{center}
	\vspace{-2mm}
\end{table*}

\section{{\ourmethod} GLUE Training Details}

For all benchmarks we used a linear rank distribution from 4 to 12 (4,5,6,6,7,8,8,9,10,10,11,12), such that the average rank is 8 (ranks rounded to integers). All eight benchmarks were trained using linear learning-rate scheduling, with the initial learning rate reported as \textit{learning rate}, and the number of epochs for the scheduler as \textit{epochs}. The runs were stopped after \textit{stop epoch} epochs. Hyper-parameters: learning rate, batch size, \# epochs, decay and prune ratio were randomly searched over the space $\{ 6\times 10^{-5},$ $1\times 10^{-4},$ $2\times 10^{-4},$ $6\times 10^{-4}, 1\times 10^{-3},$ $1.2\times 10^{-3},$ $1.5\times 10^{-3},$ $2\times 10^{-3},$ $2.3\times 10^{-3}  \},$ $\{ 4, 8, 16, 32 \},$ $\{ 10, 30, 50, 60, 70 \},$ $\{ 0, 0.1, 0.01 \},$ $ \{0.25, 0.50, 0.75 \}$ correspondingly. For all benchmarks and methods the $\textit{max seq length}$ is 128.

\begin{table*}[b]
 \vspace{1mm}
\caption{Hyper-parameters of {\ourmethod} for GLUE benchmark.}
\vspace{-1mm}
\label{tab:glue_hypers}
\begin{center}
\begin{small}
\begin{tabular}{l|ccccccc}
\toprule
{Dataset} & {learning rate} & {batch size} & {\# epochs} & stop epoch & decay & prune ratio
\\
\midrule 
{\bf MNLI} & {$1\times 10^{-4}$} & 32 & 70 & 5 & 0.01 & 0.50
\\
{\bf RTE} & $ 1.2\times 10^{-3} $ & 32 & 70 & 25 & 0.01 & 0.50
\\
{\bf QNLI}  & $ 1\times 10^{-4} $ & 32 & 60 & 3 & 0.01 & 0.50
\\
{\bf MRPC} & $ 1\times 10^{-3} $ & 32 & 60 & 15 & 0.01 & 0.50
\\
{\bf QQP } & $6\times 10^{-4}$ & 32 & 10 & 10 & 0.01 & 0.50
\\
{\bf SST-2} & $ 6\times 10^{-5} $ & 32 & 60 & 5 & 0.01 & 0.50
\\
{\bf CoLA} & $ 2\times 10^{-4} $ & 4 & 70 & 6 & 0.01 & 0.50
\\
{\bf STS-B} & $ 2.3\times 10^{-3} $ & 32 & 30 & 30 & 0.10 & 0.75
\\
\bottomrule
\end{tabular}
\end{small}
\end{center}
\end{table*}

\end{document}